# Development of a Shape-memorable Adaptive Pin Array Fixture

Peihao Shi[a], Zhengtao Hu[a], Kazuyuki Nagata[b], Weiwei Wan[a], Yukiyasu Domae[b], Kensuke Harada[ab]*.

[a]*Department of Systems Innovation, Graduate School of Engineering Science, Osaka University, Osaka, Japan;*

[b] *Artificial Intelligence Research Center, National Institute of Advanced Industrial Science and Technology, Tokyo Japan.*

*harada@sys.es.osaka-u.ac.jp

# Development of a Shape-memorable Adaptive Pin Array Fixture

This paper proposes an adaptive pin-array fixture. The key idea of this research is to use the shape-memorable mechanism of pin array to fix multiple different shaped parts with common pin configuration. The clamping area consists of a matrix of passively slid-able pins that conform themselves to the contour of the target object. Vertical motion of the pins enables the fixture to encase the profile of the object. The shape memorable mechanism is realized by the combination of the rubber bush and fixing mechanism of a pin. Several physical peg-in-hole tasks is conducted to verify the feasibility of the fixture.

## 1. Introduction

As globalization has taken hold, personalized customization of a product becomes the trend of current manufacturing. However, automation of low-volume high-mix production which is the basis of personalized customization is limited by high costs and poor efficiency.

To automate the product assembly, fixtures to fix the pose of an assembly part are crucial. However, a fixture has to be carefully designed based on the experience of those who construct a robot system performing the assembly task. In addition, a fixture has been designed for each part and has been difficult to use to fix multiple parts with different shape. Particularly in assembly tasks of low-volume high-mix production, a large number of fixtures has to be designed according to the increase of the number of parts with different shapes. Normally, the costs associated with fixtures can account for 10–20% of the total cost of a manufacturing system [1]. To cope with such a problem, we focus on an adaptive fixture which can fix multiple objects with different shape.

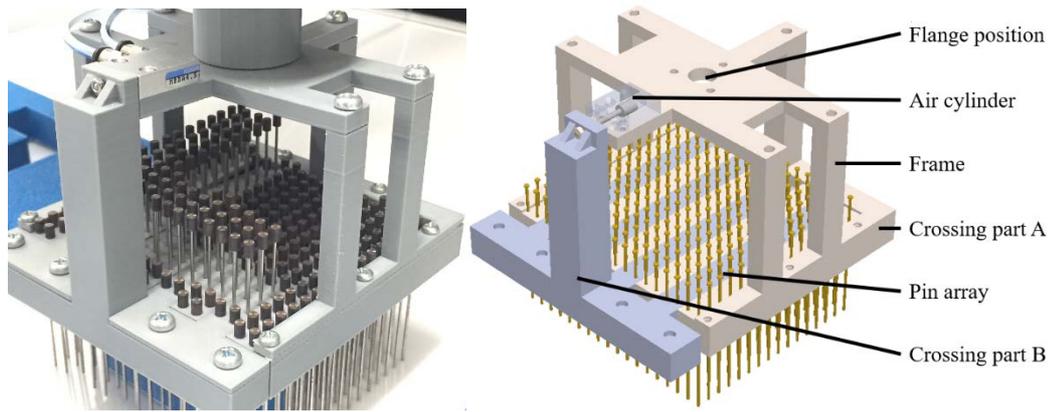

Fig. 1 The structure of the proposed fixture

We propose an adaptive fixture based on the pin array structure as shown in Fig.1. In our design, we set several pins to compose the clamping area of the fixture. The fixture can position the target part since each pin can move along vertical direction independently. The application of this fixture can increase production efficiency and reduce the cost.

With the pin array structure, the proposed fixture can fix multiple parts with different shapes. The working process of the fixture has two steps: shape-memorize and batch-clamping. In the shape-memorize step, the corresponding pins will be pushing up to memorize the shape of target part. In addition, the shape of target part will be kept in the fixture since we use the rubber rings to stop the vertical motion of these pins. Fig. 2 shows an example of the shape-memorize step. If we use the fixture to fix the triangular-shaped part (Fig. 2 (a)), the corresponding pins will be pushing up and the triangle shape will be memorized in the fixture even after we release the triangle part (Fig. 2(b)). Likewise, if we continuously use the fixture to fix the rectangle part, the rectangle shape will also be memorized by the fixture (Fig. 2 (c)). The configuration of the pins is memorized by using the rubber ring attached to each pin. Even if the two clamping area is overlapped, the two parts can independently be fixed if the form closure in 2D is satisfied in each clamping area.

On the other hand, in the batch-clamping step, we actually use the fixture for assembly task. The clamping area of our proposed fixture can be split into two crossing elements and can be moved in the horizontal direction. Due to the horizontal motion of the fixture, the assembly part can be grasped and is used for the assembly task.

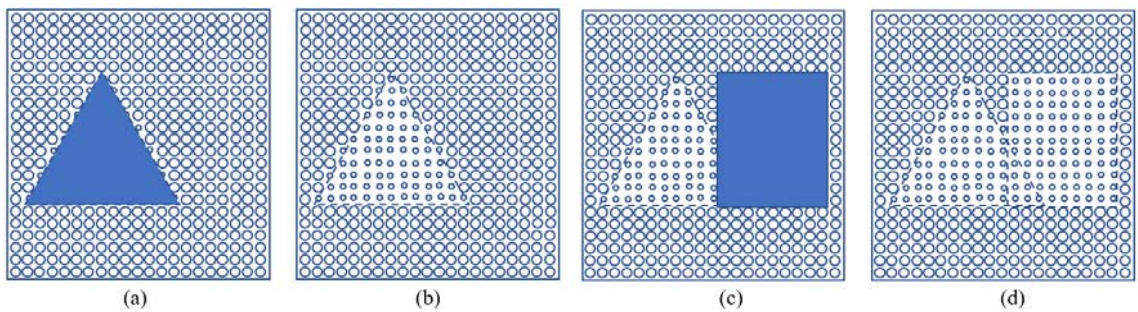

Fig.2 Shape memorize step of the pin array fixture where it can simultaneously memorize the shape of two parts with different shape.

In summary, our main contributions of this paper are as below:

1. We propose a fixture based on the pin array mechanism, which can be used for an assembly task of the high-mix manufacturing.

2. By memorizing the pin configuration, the proposed fixture can fix multiple parts with different shapes with keeping the same pin configuration.

The following is the structure of the paper: we first introduce several related works about the research on adaptive fixture design and pin array mechanism in section 2. In section 3, we show the structure of our fixture design and the detailed functions of the fixture. We construct a physical peg-in-hole task to verify the feasibility of the fixture in section 4.  Section 5 explain about the limitation of the fixture. We discuss the stability of the fixture and talk about the future works in section 6. In the last section, we propose the conclusion.

## 2. Related works

In this section, we review the related studies by separating them into two categories, i.e., a) the design of adaptive fixture, and b) application of pin array mechanism.

*2.1 The design of the adaptive fixture*

The flexible fixture is a popular research direction in the robotic and mechanical design field. Bi et.al. reviewed the studies of the flexible fixture design used in the production automation [1]. Wan et al. focused on the automatic fixture design methodology based on inheriting and reusable design [2]. De et al. proposed a reconfigurable fixture to increase the flexibility of the machining task of thin sheets parts [3]. Mu¨ller et al. presented a highly reconfigurable automated handling system to transfer the benefits of mass customization to large products with lower production quantities [4]. Olaiz et.al. proposed an adaptive fixture for accurate positioning of a planet carrier with very strict requirements of tolerances [5]. Vaughan et al. focus on improving the assembly process capability of aircraft wing in conjunction with an adaptive fixturing system [6]. Yu et al. presented a process for the control of machining distortion to eliminate surface errors and improve precision by using an adaptive dual-arm fixture [7]. Lu et al. designed a fixture used for automotive industry having high modularity and flexibility [8]. Yu et al. proposed a flexible fixture design method for similar automotive body parts to decrease the repeating fixture construction cost [9]. Do, Minh Duc et.al. proposed a novel method for optimum workpiece positioning in rod-type fixtures for thin-walled components [10]. On the other hand, this paper proposes a pin array fixture design which can be used to fix multiple parts with different shape with the same pin configuration.

*2.2 Pin array fixture mechanism*

Among the research on adaptive fixture design, pin array mechanism is a relatively new research field. Mo et al presented a pivoted pin array gripper in 2018 [11-12]. P. Fromm focuses on the gripper design based pin array mechanism [13]. Almost all those researches had focused on how to use the pin array mechanism to design a gripper. Compared with them, the originality of this paper is to propose a unique structure to fix the target part which can fix multiple parts with keeping the same pin configuration.

3. **Structure of the proposed fixture**

We explain the details of the proposed fixture in this section, including the mechanism and kinematic structure. As shown in Fig. 3, the fixture is composed of multiple pins to memorize the shape of the target object and the rubber bush attached to each pin. The fixture includes two crossing elements (Fig. 3 (a)) where the clamping motion can be generated by the relative motion between these two crossing elements.

*3.1 Shape-memorizing mechanism*

We assume that the target part is stably placed on the horizontally flat table and that its pose does not change even if the external force is applied from the vertically upward direction. In the shape-memorizing step, the fixture pushes the target object from the vertically upward direction. After the tip of pins contact the target object, vertical motion of the pins enables the fixture to encase the profile of the target part. A rubber ring is attached to each pin as shown in Fig. 3 (b). A rubber ring is used to temporary fix the pins' configuration and to memorize the shape of target object by the effect of friction between a rubber ring and a pin. However, the friction force of a rubber ring is not strong enough to keep a pin's configuration when the external force is applied to a pin. To fix the pins' configuration under the external force applied to the part, the fixture includes the fixing mechanism of pins as shown in Fig. 3 (c). If we press a rubber ring from the top, the bottom part of the rubber ring is wedged. Due to the wedging force, the pin's configuration is fixed. On the other hand, to release a pin, we press the rubber ring from the bottom.

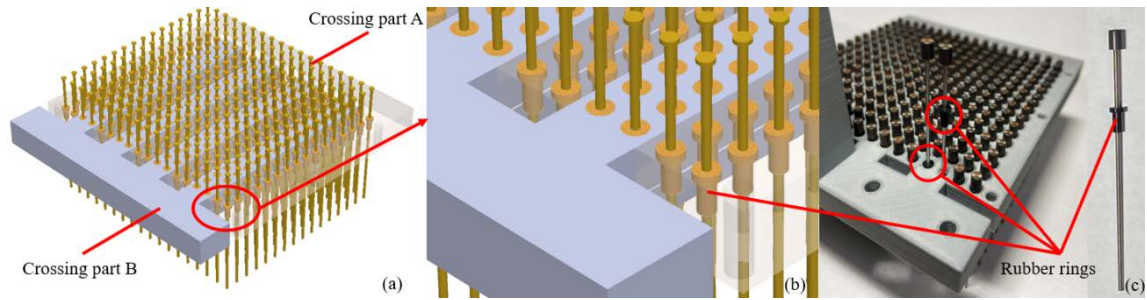

Fig. 3 Rubber rings in the fixture

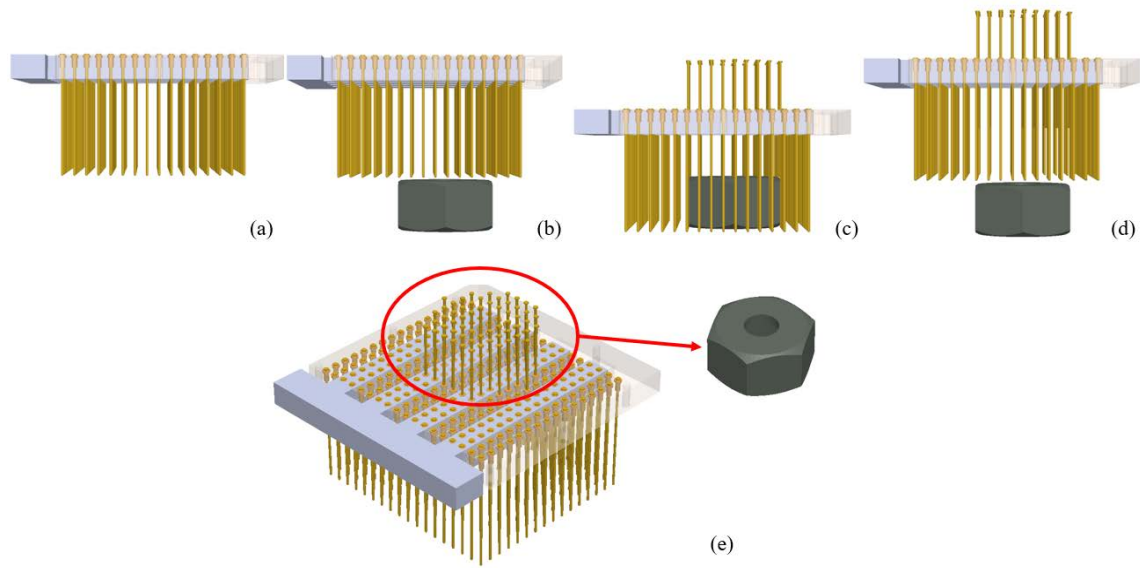

Fig. 4 The pins are pushed up by target object and keep their position because of the friction of rubber rings

Fig. 4 depicts the working principle of the fixture. The corresponding pins are pushed up by the target part (nut). Then, the pins' configuration is kept because of the friction of rubber rings. The shape of target object is memorized by the configuration of pin array as shown in Fig.4 (e). Likewise, we can use the different areas of the pin array to memorize the shape of other parts. After memorizing the shape of multiple parts, we fix the pins' configuration by using the fixing mechanism. Then, it can be used for assembly tasks.

Here, we note that, if the pins are finely distributed, the fixture can be precisely memorized the shape of the target object. On the other hand, according to the distance between two neighbouring pins, the gap between the contour of target part and a pin closest to the contour occurs.

*3.2 Clamping motion*

After memorizing the shape of multiple parts in the shape-memorizing step, we use the fixture to actually fix one of the parts used for an assembly task. This subsection depicts how the fixture clamps the target part in the batch-clamping step.

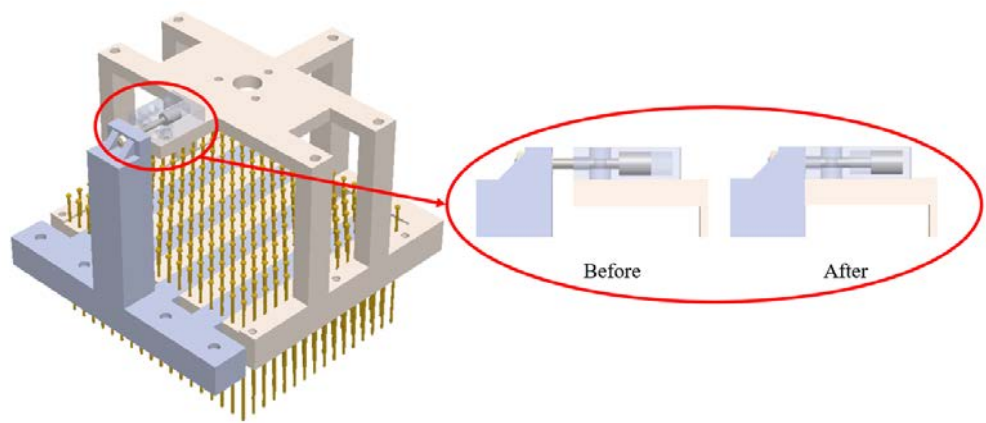

Fig. 5 The relative motion of the crossing parts

In this step, the fixture clamps one of the parts which shape is memorized in the shape-memorizing step. After placing the fixture such that the target part is located at its corresponding clamping area of pin array, the fixture tries to clamp the target part. The relative motion between two crossing elements in the horizontal direction provides the clamping force of pins in the horizontal direction to fix the target part. The actuator used to generate the horizontal motion of the two crossing elements is shown in Fig.5.

We analyse the mechanism where the target part can be fixed by the clamping motion of the crossing elements. We assume that, after the target object is clamped, the friction force in the vertical direction between the target part and a pin is large enough. We check if the form closure in the 2D horizontal plane is satisfied by the clamping motion of the crossing elements.

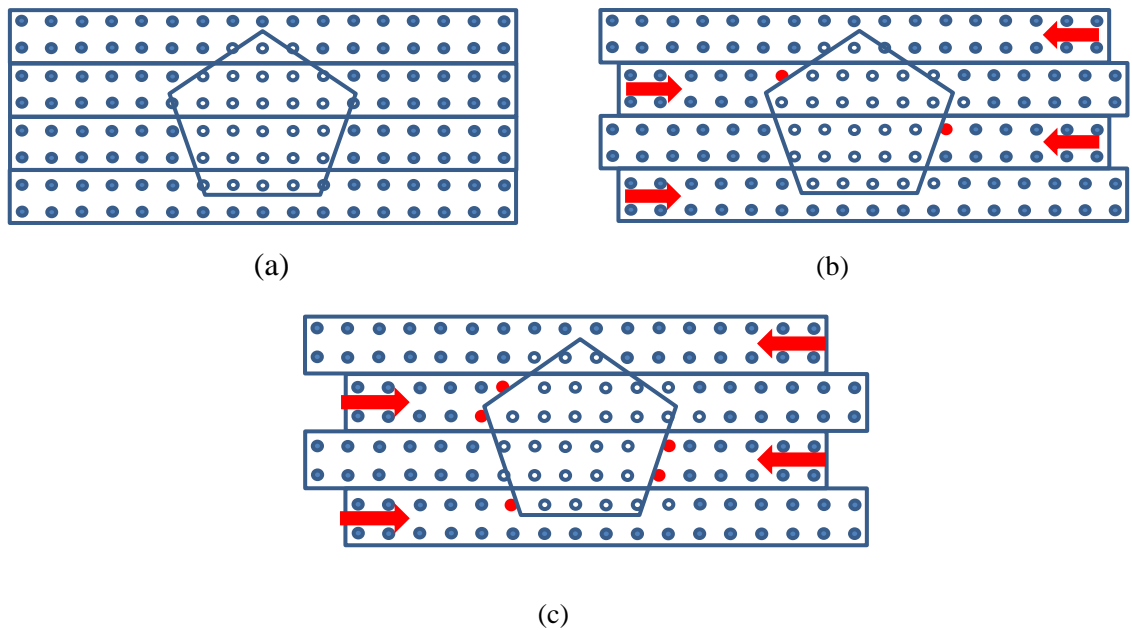

Fig.6 The clamping motion of the fixture where (a) shows the pin configuration after the shape-memorizing step, the hollow circles indicate the stuck pins, (b) shows the situation in which two pins enter in contact with the target part, and (c) shows the situation in which the form closure is satisfied.

In the shape-memorizing step, by the motion of the fixture compressing an object in the vertical direction, the pins contacting the object are displaced in the vertical direction. However, the pins without displaced do not exert any forces to the target part in the horizontal direction at this time (Fig.6(a)). In the batch-clamping step, by the clamping motion of the crossing elements in the horizontal direction, two pins near the clamping area enter in contact with the target part as shown in Fig.6(b). In this state, the form closure is not satisfied since four contacts are necessary to satisfy the form closure in the 2D plane. From this state, the crossing elements is further displaced. The pins which have already contacted the target part cause elastic displacement and keep contacting with the target part. Finally, the form closure is satisfied after more than four contacts are established.

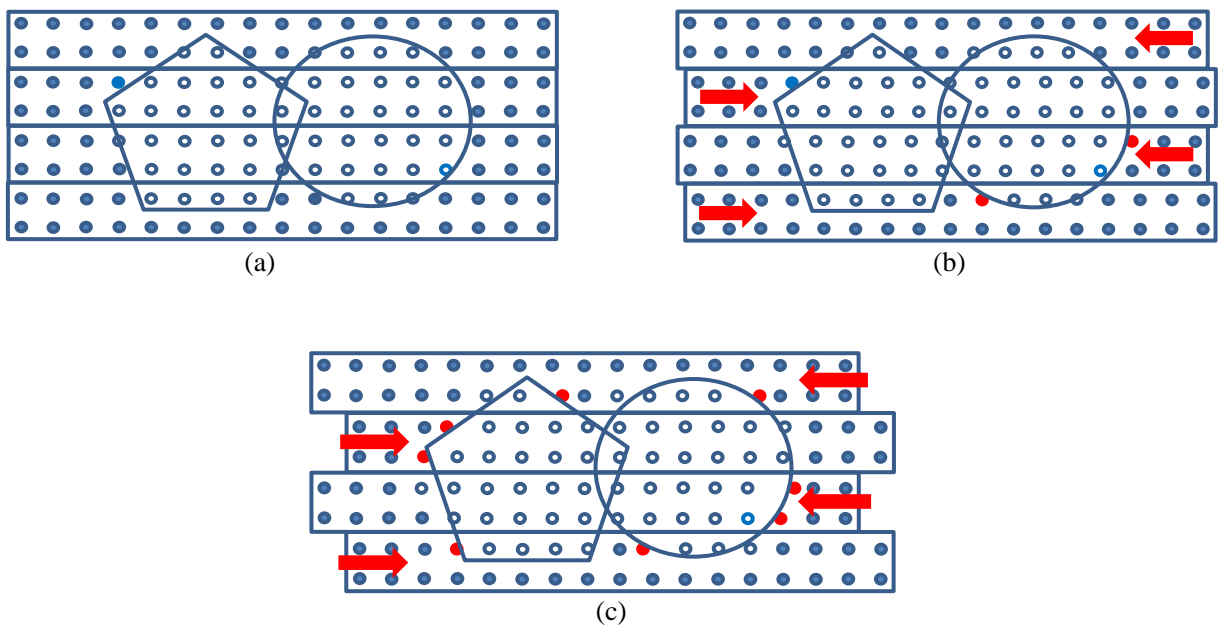

Fig.7 The clamping motion for two target parts where (a) shows the pin configuration after the shape-memorizing step, (b) shows the situation in which two pins enter in contact with the target parts, and (c) shows the situation in which the form closure is satisfied for both parts.

If we memorize the shape of multiple parts in the shape-memorizing step, an overlapping area between two parts is encountered as shown in Fig. 7(a). With the clamping motion of the crossing elements, the pins included in the overlapping area will not contact the target object. However, even in such cases, the fixture can fix the target part if the 2D form closure can be satisfied by using the remaining contacts with the pins as shown in Fig. 7(c).

Here, we note that, if the pins are finely distributed, the amount of relative motion between two crossing elements can be very small. In this case, the pose of the target object can be precisely defined and the amount of the pins' elastic displacement can be small. On the other hand, if the distance between two neighbouring pins becomes larger, the pose of clamped target part becomes different from the pose of target object in the shape-memorizing step. In this case, the pose of target object is determined by minimizing the pins' elastic energy.

Let $\boldsymbol{n}_i = (x_i, y_i)^T$ be the unit inner normal vector at the $i$-th ($i=1,\cdots,c$) contact point of the object projected onto the horizontal plane. Let $\boldsymbol{V} = (\boldsymbol{v}_{obj}{}^T, w_{obj})^T \in R^3$ be the linear and angular velocity of the target part. The condition for the form closure can be satisfied when the solution of the following equation

$$N^T G^T V \geq 0 \tag{1}$$

is $\boldsymbol{V}=0$, where

$$N = \mathrm{diag}(\boldsymbol{n}_1, \cdots, \boldsymbol{n}_c), \tag{2}$$

$$G = \begin{bmatrix} \boldsymbol{I} & \cdots & \boldsymbol{I} \\ [\boldsymbol{p}_1 \times] & \cdots & [\boldsymbol{p}_c \times] \end{bmatrix}, \tag{3}$$

and $\boldsymbol{p}_i$ denotes the position vector of the $i$-th contact point. This condition can be examined by solving the linear programming problem. But, one of the necessary conditions is that the matrix $GN$ is full row rank [14-15].

## 4. Experiment

We construct the proposed fixture and connect it to the wrist of UR-3 robot as shown in

Fig. 8. Peg-in-hole tasks of parts with several different shapes are conducted to verify the feasibility of the proposed fixture. We choose the peg-in-hole of alphabet parts as the target.

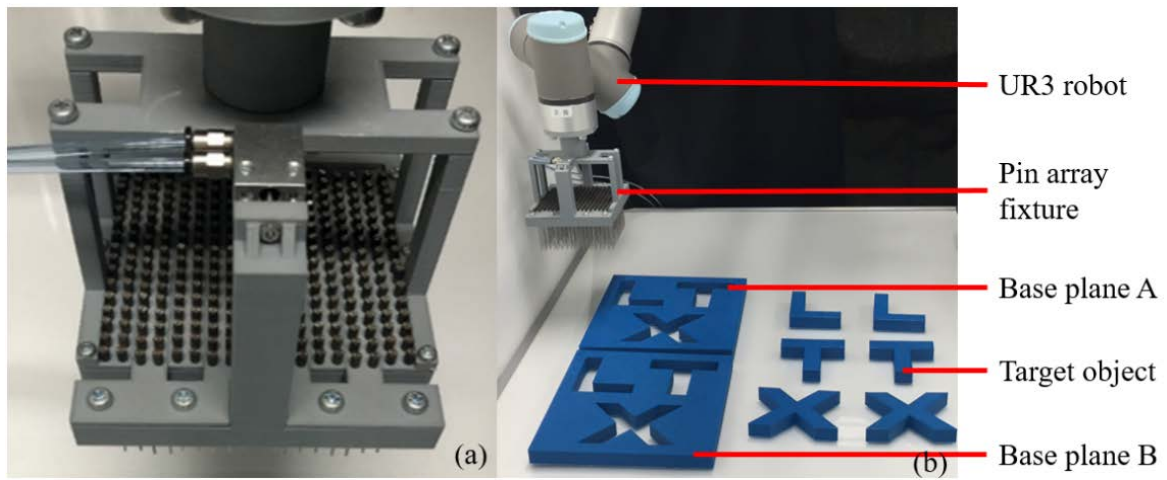

Fig. 8 Experimental setup

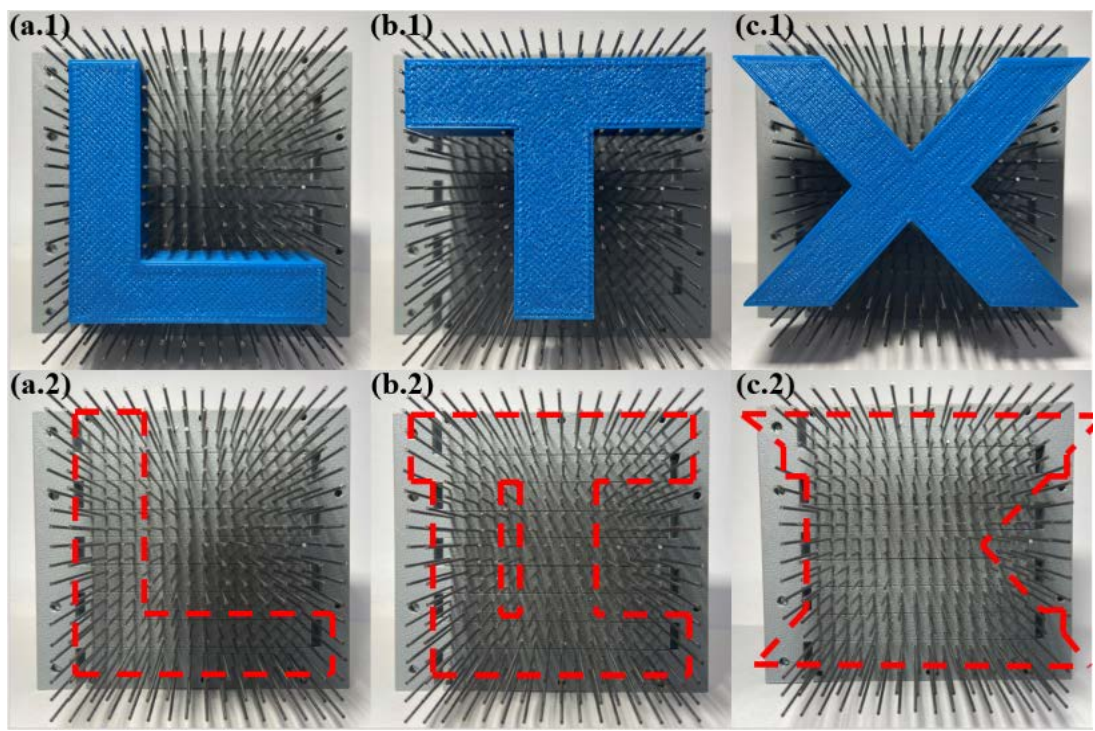

Fig. 9 The fixture memorized the "L", "T", "X" shape after fix the first group of part

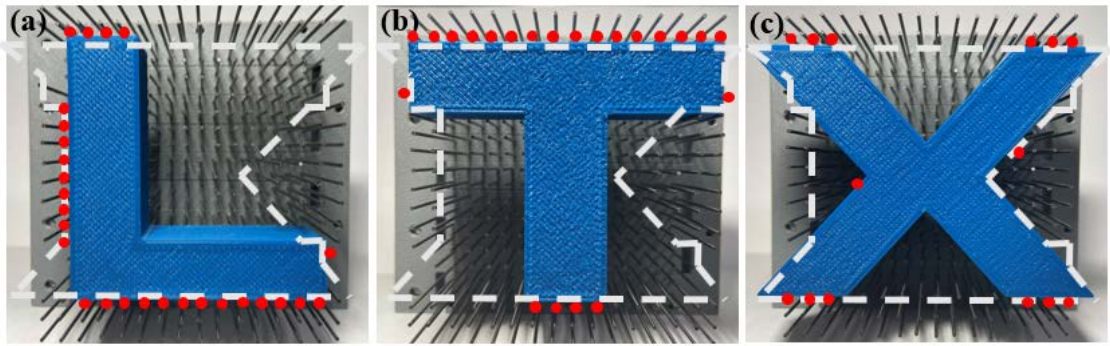

Fig. 10 The batch parts with the same shape can be fixed after the fixture memorized the shape. (Contacting pins are marked in red)

Fig. 9 shows the pin array of the fixture after memorized "L", "T", "X" in order where the red dashed line shows the clamped area. In the batch-clamping phase, the form closure can be satisfied and we can fix these three parts by using the common pin configuration as shown in Fig. 10 where the contacting pins are marked in red. The snapshot of peg-in-hole experiment of these parts is shown in Fig. 11.

We further conducted the peg-in-hole experiment by using three groups of target parts: F/T, G/C, big and small circle parts as shown in Fig. 12. These are the examples where we need to carefully choose the clamping area.

Fig.13 shows a successful case of fixing "F" and "T" parts. The fixture can memorize the shape of both "F" (a.2) and "T" (b.2) parts at the same time. In this case, the experiment was succeed in placing both "F" and "T" parts since the vertical lines between "F" and "T" does not share the same clamping area as depicted in this figure.

On the other hand, Fig 14 shows the failure case. The pick-and-place experiment failed since the vertical lines between "F" and "T" shares the same clamping area as depicted in this figure and the form closure can be satisfied for clamping neither "F" nor "T".

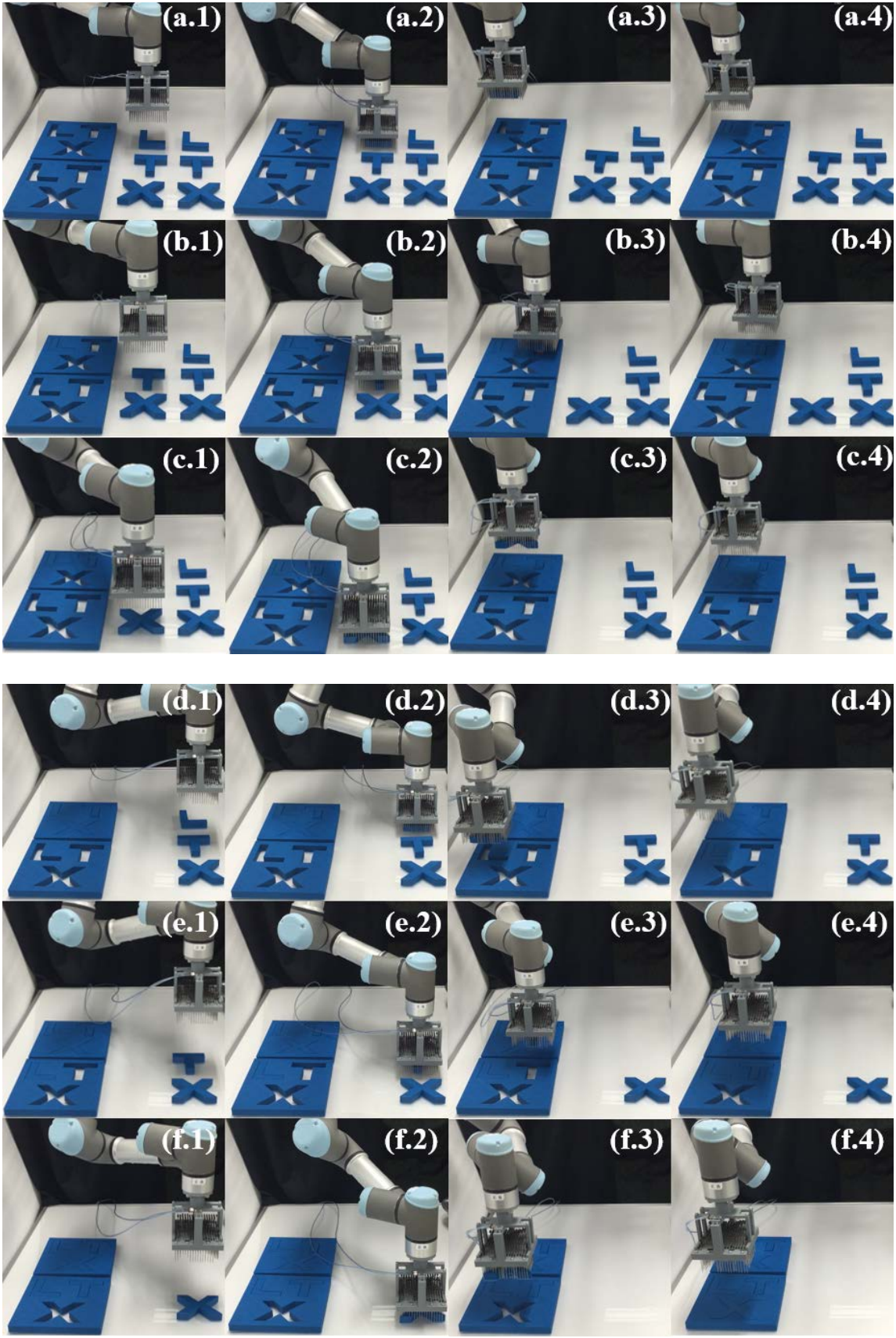

Fig. 11 Result of the peg in hole experiment

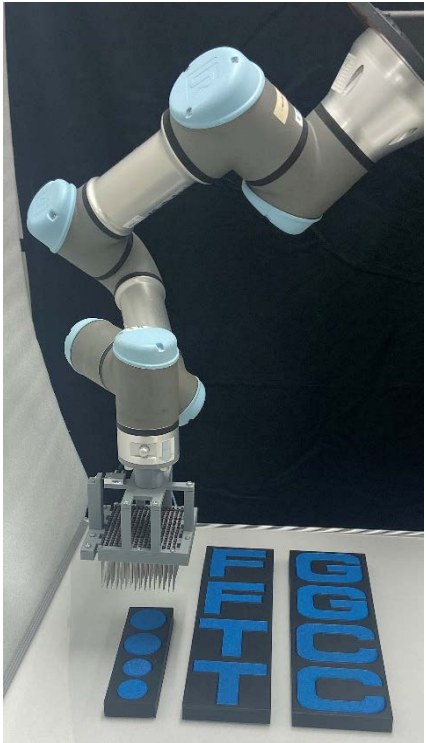

Fig. 12 The fixture can't align the target part in some cases

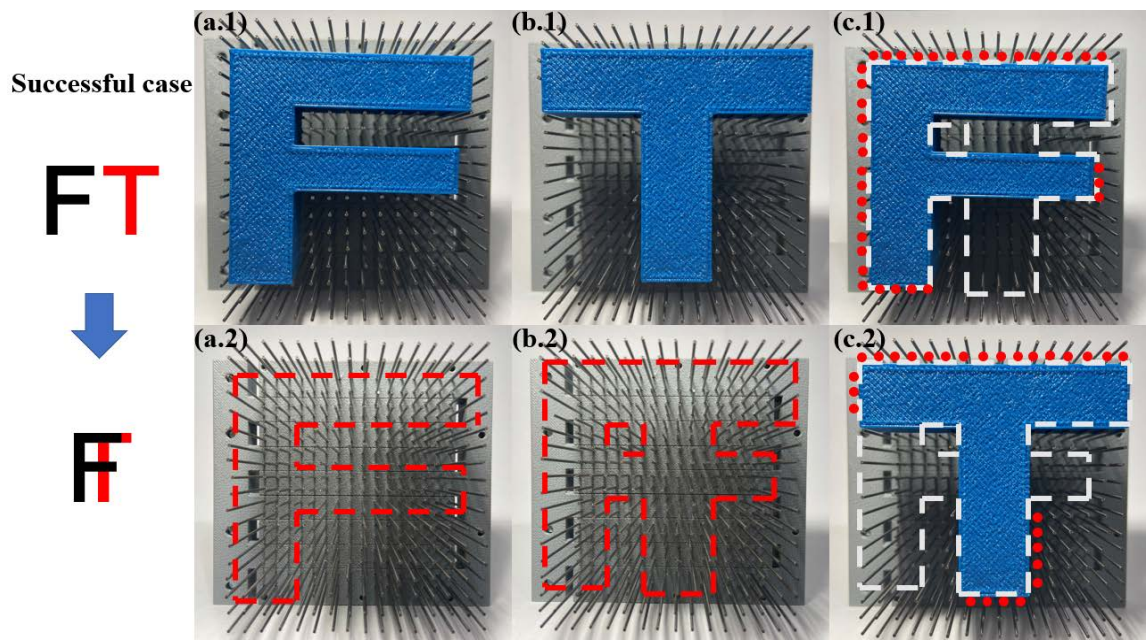

Fig.13 Pin array by which we can fix both "F" and "T" parts where contacting pins are marked in red

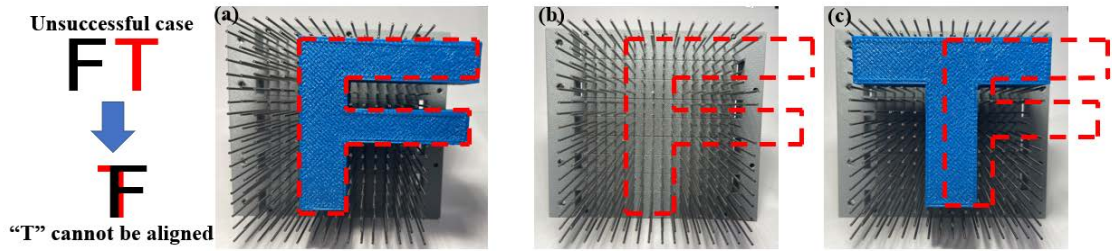

Fig. 14 Pin array by which we can fix neither "F" nor "T" parts

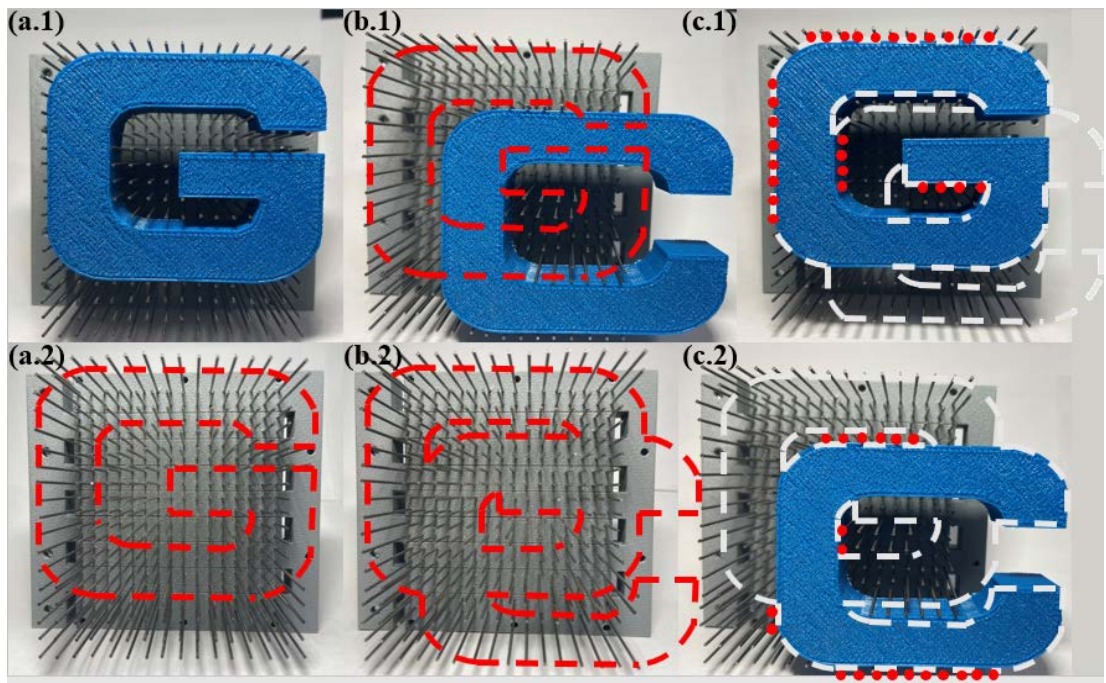

Fig. 15 Pin array by which we can fix both "G" and "C" parts

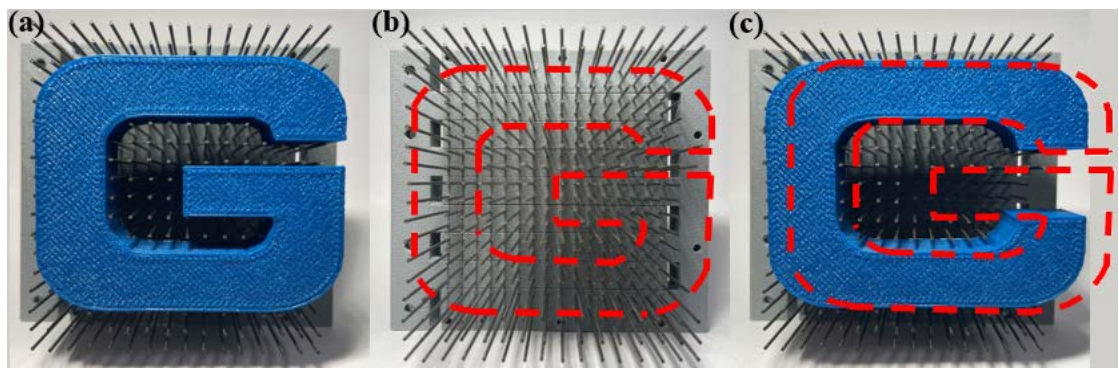

Fig. 16 Pin array by which we can fix neither "G" nor "C" parts

Fig.15 shows the successful case of fixing both "G" and "C" parts. On the other hand, Fig. 16 shows the failure case where the form closure can be satisfied for clamping neither "G" nor "C" parts. Fig.17 shows the successful case of fixing two cylinders with different radius. The experiment was successful since the shape of two cylinders are memorized in the different area of pin array. On the other hand, Fig. 18 shows the failure case of fixing a cylinder with smaller radius. Experiment was failed since clamped area of the cylinder with smaller radius is included in the clamped area of the cylinder with larger radius.

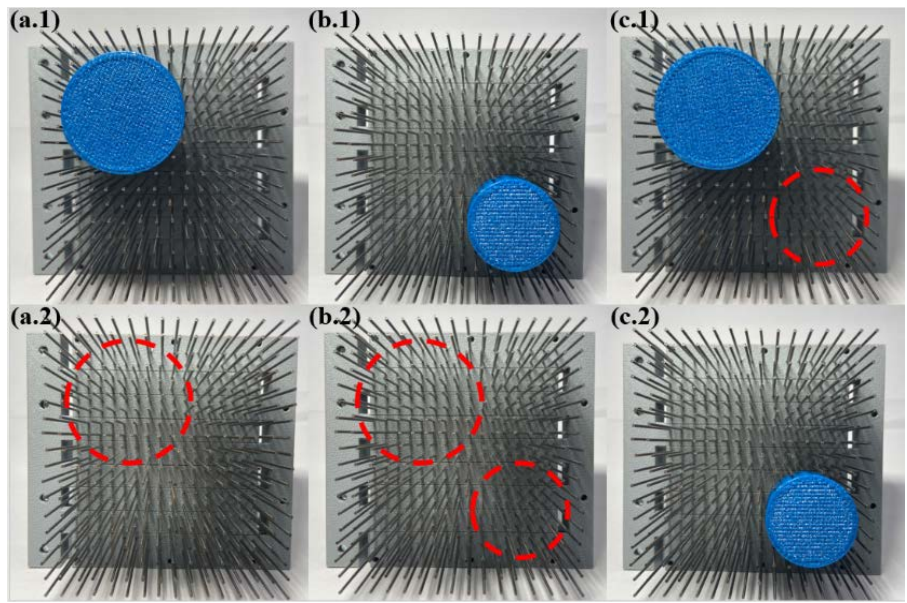

Fig. 17 Pin array by which we can fix two cylinders with different radius

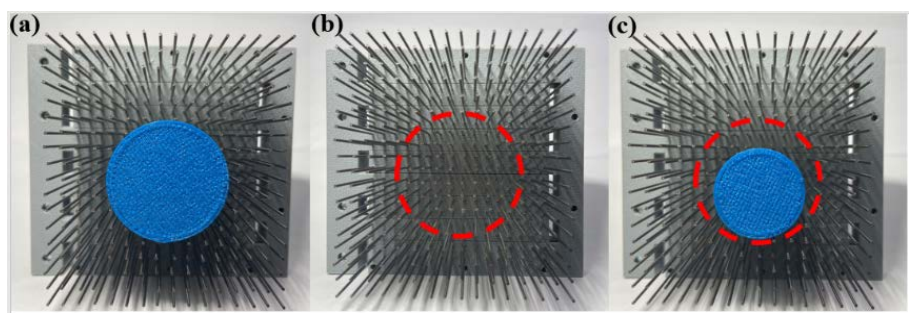

Fig. 18 Pin array by which we cannot fix a cylinder with smaller radius

## 5. Discussion

In this section, we discuss how the fixture well fixes multiple parts with different shape by using the common ping configuration. As a necessary condition for the form closure, we checked the rank of the matrix **GN** in eq. (1) as a function of the distance between two memorised areas. We use two target parts, i.e., an equilateral triangular prism and a square pillar.

The pin is simplified as a point, and the distance between two pins is 5 mm as shown in Fig.19(a). In the first case, an equilateral triangular prism is used as the target part where the dimension of the bottom surface is width w = 45 [mm] and the height h= 45 [mm]. As shown in Fig.19 (b), the triangular prism was fixed with 17 contact points. The matrix GN is full row rank in this case. In the second case, a square pillar is used as the target part where the edge length of the square is w = 45 [mm]. It should be noticed that if there is no overlap between two memorised areas, the form closure is satisfied [16].

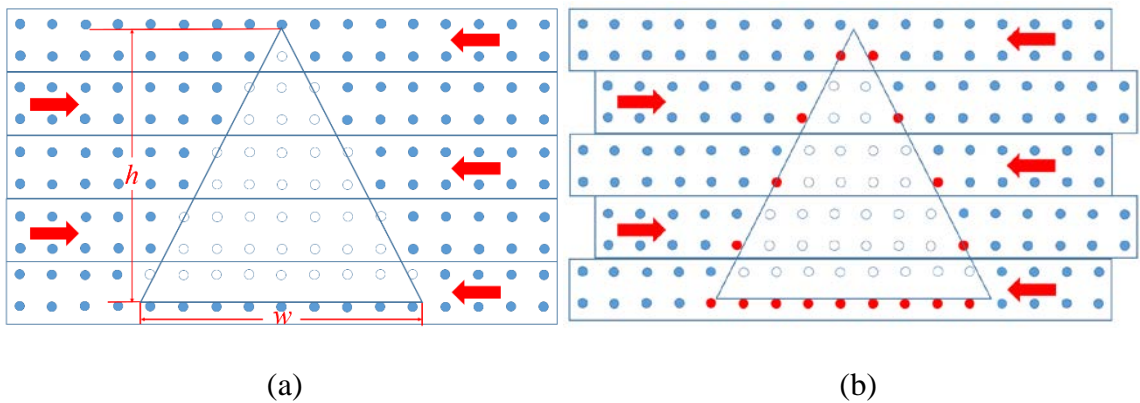

(a)                                                               (b)

Fig. 19 Fixing an equilateral triangular prism by using the proposed fixture

To fix a square pillar, the number of the contact points becomes 24 and the matrix GN becomes full row rank (Fig. 20). Then, we consider a situation where there is overlap between two memorised areas.

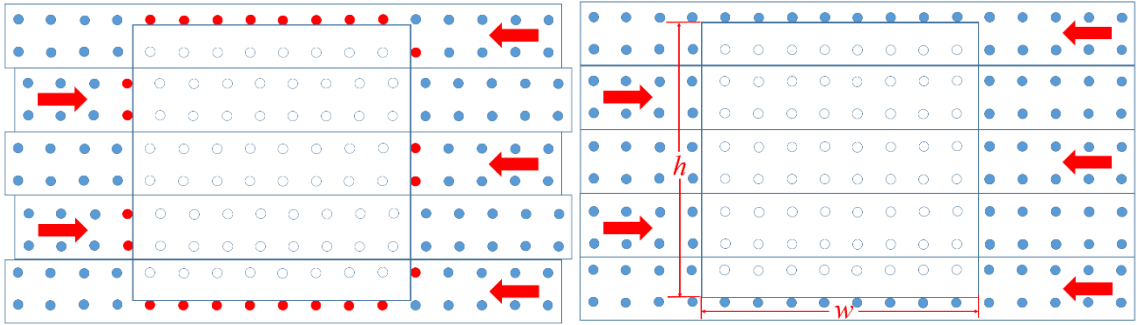

Fig.20 Fixing a square pillar by using the proposed fixture

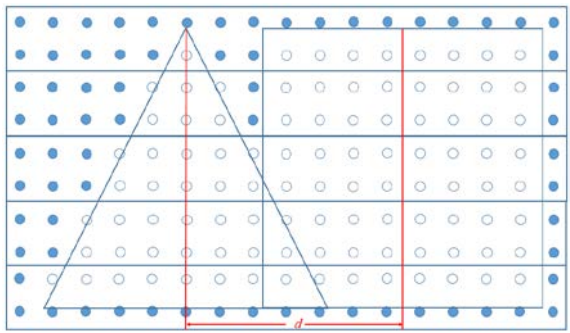

Fig.21 The distance between two memorized areas

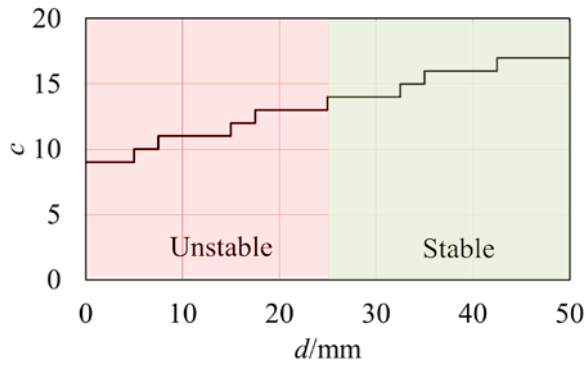

Fig.22 The form closure condition when memorizing two objects with different shape

Let $d$ and $c$ be the distance between the two memorised areas (Fig. 21) and the number of the contact points on the triangular prism. Fig.22 shows the relationship between $c$ and $d$. It can be seen that the number of the contact points increases as $d$ increases. When $d$ is larger than 250 [mm], the necessary condition for form-closure is satisfied.

**Conclusion**

This paper proposes a pin array fixture for high-mix production. It can be used to fix multiple parts with different shapes. We also can fix the batch parts with the same shape after the fixture memorized the shape. The results of peg-in-hole assembly experiments verify the feasibility of the fixture. However, there are still several deficiencies that need to improve on the proposed design. We will optimize the design to avoid the limitation in the future.

References

1. Z. Bi, W. Zhang, Flexible fixture design and automation: review, issues and future directions, International Journal of Production Research 39 (13) (2001) 2867–28

2. Wan N, Wang Z and Mo R. An intelligent fixture design method based on smart modular fixture unit. Int J Adv Manuf Tech 2013; 69: 2629–2649.

3. De LL, Zoppi M and Xiong L. A multi-robot-based reconfigurable fixture. Ind Robot 2013; 40: 320–328.

4. Mu¨ller R, Esser M and Vette M. Reconfigurable handling systems as an enabler for large components in mass customized production. J Intell Manuf 2013; 24: 977–990.

5. Olaiz, Edurne, et al. Adaptive fixturing system for the smart and flexible positioning of large volume workpieces in the wind-power sector. Procedia CIRP 21 (2014): 183-188.

6. Vaughan D, Branson D, Bakker OJ, et al. Towards self-adaptive fixturing systems for aircraft wing assembly. SAE technical paper 2015-01-2493, 2015

7. Yu JH, Chen ZT, Jiang ZP, et al. A control process for machining distortion by using an adaptive dual-sphere fixture. Int J Adv Manuf Tech 2016; 86: 3463–3470.

8. Lu S, AhmadZ, Zoppi M, et al. Design and testing of a highly reconfigurable fixture with lockable robotic arms. J Mech Des 2016; 138: 085001–085009.

9. Yu K, Wang S, Wang Y, et al. A flexible fixture design method research for similar automotive body parts of different automobiles[J]. Advances in Mechanical Engineering, 2018, 10(2): 1687814018761272.

10. DO, Minh Duc; SON, Younghoon; CHOI, Hae-Jin. Optimal workpiece positioning in flexible fixtures for thin-walled components. *Computer-Aided Design*, 2018, 95: 14-23.

11. Mo, A., Zhang, W.: Pin array hand: a universal robot gripper with pins of ellipse contour. In: 2017 IEEE International Conference on Robotics and Biomimetics (ROBIO), pp. 2075–2080. IEEE (2017)

12. Mo A, Fu H, Zhang W. A universal gripper base on pivoted pin array with chasing tip[C], International Conference on Intelligent Robotics and Applications. Springer, Cham, 2018: 100-111.

13. Fromm P M, Bradway J J, Ruiz E, et al. Air pressure loaded membrane and pin array gripper: U.S. Patent 9,925,799[P]. 2018-3-27.
14. Bicchi A. On the closure properties of robotic grasping[J]. The International Journal of Robotics Research, 1995, 14(4): 319-334
15. Caihua X, Han D, Youlun D. Fundamentals of robotic grasping and fixturing[M]. World Scientific, 2007.
16. Markenscoff, Xanthippi, Luqun Ni, and Christos H. Papadimitriou. The geometry of grasping. The International Journal of Robotics Research 9.1 (1990): 61-74.